\documentclass[onecolumn]{IEEEtran} 
\usepackage{cite}

\usepackage{color} 
\usepackage[cmex10]{amsmath}
\usepackage{algorithm}
\usepackage{algorithmic}
\usepackage{hyperref}
\usepackage[tight,footnotesize]{subfigure}
\usepackage{multirow}
\usepackage{booktabs}

\newtheorem{theorem}{Theorem}

\newtheorem{definition}{Definition}
\newtheorem{proposition}{Proposition}

\begin{document}

\title{A Theoretical Assessment of Solution Quality in  Evolutionary  Algorithms for the   Knapsack Problem}

\author{Jun He, Boris Mitavskiy and Yuren Zhou 
\thanks{Jun He and Boris Mitavskiy are with Department of Computer Science, Aberystwyth University, Aberystwyth SY23 3DB, U.K.}
\thanks{Yuren Zhou is with School of Computer Science and Engineering, South China University of Technology, Guangzhou 510640, China.}}

\maketitle

\begin{abstract} Evolutionary algorithms are well suited for solving the knapsack problem. Some empirical studies claim that evolutionary algorithms can produce good solutions to the 0-1 knapsack problem. Nonetheless, few rigorous investigations address the quality of solutions that evolutionary algorithms may produce for the knapsack problem. The current paper focuses on a theoretical investigation of three types of (N+1) evolutionary algorithms that exploit bitwise mutation, truncation selection, plus different repair methods for the 0-1 knapsack problem. It assesses the solution quality in terms of the approximation ratio. Our work indicates that the solution produced by pure strategy and mixed strategy evolutionary algorithms is arbitrarily bad. Nevertheless,  the   evolutionary algorithm using helper objectives may produce 1/2-approximation solutions to the 0-1 knapsack problem.
 \end{abstract}

\section{Introduction}
The knapsack problem is an NP-hard combinatorial optimisation problem~\cite{martello1990knapsack}, which includes a variety of knapsack-type problems such as the 0-1 knapsack problem and multi-dimensional knapsack problem. In the last two decades, evolutionary algorithms (EAs), especially genetic algorithms (GAs), have been well-adopted for tackling the knapsack problem \cite{michalewicz1994genetic,khuri1994zero,chu1998genetic,raidl1998improved}.
The problem has received a particular interest from the evolutionary computation community for the following two reasons. The first reason is that the binary vector representation of the candidate solutions is a natural encoding of the 0-1 knapsack problem's search space. Thereby, it provides an ideal setting for the applications of genetic algorithms~\cite{michalewicz1996genetic}. On the other hand, the multi-dimensional knapsack problem is a natural multi-objective optimization problem, so that it is often taken as a test problem for studying multi-objective optimization evolutionary algorithms (MOEAs)~\cite{zitzler1999multiobjective,jaszkiewicz2002performance,eugenia2003solving,ozcan2009case,kumar2010assessing}.

A number of empirical results in the literature (see, for instance,   \cite{zitzler1999multiobjective,jaszkiewicz2002performance,
eugenia2003solving,ozcan2009case,kumar2010assessing,rohlfshagen2010nature}) assert that EAs can produce ``good'' solutions to the knapsack problem. A naturally arising question is then how to measure the ``goodness'' of solutions that EAs may produce? To address the question, the most popular approach is to compare the quality of the solutions generated   by   EAs  via computer experiments. For example, the solution quality of an EA  is measured by the best solution
found within 500 generations~\cite{michalewicz1996genetic}.
Such a comparison may help to compare performance of different EAs, yet it seldom provides any information regarding the proximity of the solutions produced by the EAs to the optimum.

From the viewpoint of algorithm analysis, it is important to assess how ``good'' a solution is in terms of the notion of approximation ratio (see \cite{williamson2011design}). There are several effective approximation algorithms for solving the knapsack problem \cite{martello1990knapsack}. For example, a fully polynomial time approximation scheme for the 0-1 knapsack problem has been presented in \cite{ibarra1975fast}. Nonetheless, very few rigorous investigations addressing the approximation ratio of EAs on the 0-1 knapsack problem exist. \cite{kumar2006analysis} recast the 0-1 knapsack problem  into a bi-objective knapsack problem   with two conflicting objectives (maximizing profits and minimizing weights). A (1+$\epsilon$)-approximate set of the knapsack problem has been introduced for the bi-objective optimization problem. An MOEA, called   Restricted Evolutionary Multiobjective Optimizer, has been designed to obtain the (1+$\epsilon$)-approximate set. A pioneering contribution of \cite{kumar2006analysis} is a rigorous runtime analysis of the proposed MOEA.

The current paper focuses on investigating the approximation ratio of three types of $(N+1)$ EAs combining bitwise mutation, truncation section and diverse repair mechanisms for the 0-1 knapsack problem. The first type is several  pure strategy EAs, where a single repair method is exploited in the EAs. The second type is several mixed strategy  EAs,  which choose a repair method from a repair method pool randomly.   The third type is a multi-objective  EA using helper objectives, which is a simplified version of the EA in~\cite{he2014novel}. 

The remainder of the paper is organized as follows. The 0-1 knapsack problem is introduced in section \ref{secProblem}. In  section~\ref{secPure} we analyse pure strategy   EAs, while in section~\ref{secMixed} we analyse mixed strategy   EAs. Section~\ref{secMOEA} is devoted to analysing an  MOEA using helper objectives. Section~\ref{secConclusions} concludes the article.

%%%%%%%%%%%%%%%%%%%%%%%%%%%%%%%%%%%%%

%%%%%%%%%%%%%%%%%%%%%%%%%%%%%%%%%%%%%
\section{Knapsack Problem and Approximation Solution} \label{secProblem}
The 0-1 knapsack problem is the most important knapsack problem and one of the most intensively studied  combinatorial optimisation problems \cite{martello1990knapsack}. Given an instance of the 0-1 knapsack problem with a set of weights
$w_i$, profits $p_i$, and capacity $C$ of a knapsack, the task is to find a binary vector $\vec{x}=(x_1 \cdots x_n)$ so as to
\begin{equation}
\label{equ:single-objective}
\begin{array}{llll}
&\max_{\vec{x}}  \sum^n_{i=1} p_i x_i,  \\
&\mbox{subject to } \sum^n_{i=1} w_i x_i \le C,
\end{array}
\end{equation}
where  $
x_i =   1$ if the item $i$ is
selected in the knapsack and $x_i=
0$ if the item $i$ is not selected in the knapsack. 
A feasible solution is a knapsack represented by a binary vector $\vec{x} = (x_1   x_2  \cdots  x_n)$ which satisfies the constraint. An infeasible one is an $\vec{x}$ that violates the constraint. The vector $(0 \cdots 0)$  represents a null knapsack.

In   last two decades, evolutionary algorithms, especially genetic algorithms (GAs), have been well adopted for tackling the knapsack problem~\cite{michalewicz1994genetic,khuri1994zero}. In order to assess the quality of solutions in EAs, we follow the classical $\alpha$-approximation algorithm (see \cite{williamson2011design} for a detailed exposition) and define an evolutionary approximation algorithm as follows. 

\begin{definition}\label{alphaApproxDefn}
We say that an EA is an $\alpha$-approximation algorithm for an optimization problem if for all instances of the problem, the EA can produce a solution within a polynomial runtime, the value of which is within a factor of $\alpha$ of the value of an optimal solution, regardless of the initialization. Here the runtime is measured by the expected number of function evaluations.

\end{definition}

For instance, in case of the 0-1 knapsack problem, an evolutionary $1/2$-approximation algorithm  always can find a solution the value of which is at least a half of the optimal value within a polynomial runtime.

\section{Pure Strategy (N+1) Evolutionary Algorithms}
\label{secPure}
In this section we analyze pure strategy $(N+1)$ EAs  for the 0-1 knapsack problem. Here a \emph{pure strategy EA} refers to an EA that employs a single repair method.
  The genetic operators used in  $(N+1)$ EAs are
bitwise mutation   and  truncation
selection.  
\begin{itemize}
\item \emph{Bitwise Mutation:}  Flip each bit  with probability $ {1}/{n}$.
\item \emph{Truncation Selection:} Select the best N individuals from the parent population and the child. 
\end{itemize}

A number of diverse methods are available to handle constraints in EAs~\cite{michalewicz1996genetic,coello2002theoretical}. Empirical results indicate that repair methods are more efficient than  penalty function methods for the  knapsack
problem~\cite{he2007comparison}. Thus, only the repair methods are investigated in the current paper. The repair procedure \cite{michalewicz1996genetic} is explained as follows.
\begin{algorithmic}[1]
\STATE \textbf{input} $\vec{x}$;
\IF{$\sum^n_{i=1} x_i w_i >C$}
\STATE $\vec{x}$ is infeasible;
\WHILE{($\vec{x}$ is infeasible)}
\STATE $i=:$ \textbf{select} an item from the knapsack;
\STATE set $ x_i=0$;
\IF{$\sum^n_{i=1} x_i w_i \le C$}
\STATE $\vec{x}$ is feasible;
\ENDIF
\ENDWHILE
\ENDIF
\STATE \textbf{output} $\vec{x}$.
\end{algorithmic}

There are several \textbf{select} methods available for the \textbf{repair} procedure, such as the \emph{profit-greedy repair}, the \emph{ratio-greedy repair} and the \emph{random repair methods}.
\begin{enumerate}
\item
\emph{Profit-greedy repair:}
sort the items $x_i$ according to the decreasing order of their corresponding profits $p_i$. Then select the item with the smallest profit and remove it from the knapsack.
\item
\emph{Ratio-greedy repair:}
sort the items $x_i$ according to the decreasing order of the corresponding ratios ${p_i}/{w_i}$. Then select the item with the smallest ratio and remove it from the knapsack.

\item
\emph{Random repair:} select an item $x_i$ from the knapsack at random and remove it from the knapsack.
\end{enumerate}
Thanks to the repair method, all of the infeasible solutions have been repaired into the feasible ones.
The fitness function of a feasible solution $\vec{x}$ is $
  f(\vec{x}).
$

First, let's consider a pure strategy $(N+1)$ EA using ratio-greedy repair for solving the 0-1 knapsack problem, which  is described  as follows.  
 
 \begin{algorithmic} [1]
\STATE \textbf{input} an instance of the 0-1 knapsack problem;
\STATE initialize a population considering of N individuals;
\FOR{$t=0, 1,2, \cdots$}
\STATE    mutate  one individual and generate a child;
\IF{the child  is an infeasible solution}
\STATE repair it into a feasible solution using the ratio-greedy repair;
\ENDIF
 
\STATE  select N individuals from the parent population and the child using   truncation selection;
 
\ENDFOR
\STATE \textbf{output} the maximum of the fitness function.
\end{algorithmic}

The following proposition reveals that the $(N+1)$ EA using the ratio-greedy repair cannot produce a good solution to the 0-1 knapsack problem within a polynomial runtime.
\begin{proposition}
\label{thmRatioRepair}
For any constant $\alpha \in (0,1)$, the $(N+1)$ EA using Ratio-Greedy Repair is not an $\alpha$-approximation algorithm for the 0-1 knapsack problem.
\end{proposition}
\begin{IEEEproof}
According to definition~\ref{alphaApproxDefn}, it suffices to consider the following instance of the 0-1 knapsack problem:
\begin{center}
\begin{tabular}{c|c|c |c}
\toprule
Item $i$ & $1$ &   $2, \cdots,  \alpha n $ &$ \alpha n  +1, \cdots, n$\\
\midrule
Profit $p_i$ &  $n$ &$ 1$ &$ \frac{1}{n}$\\
Weight $w_i$ &  $n$  &  $\frac{1}{\alpha n}  $  &$ n$   \\
\midrule
Capacity &\multicolumn{3}{c}{$n  $}  \\
\bottomrule
\end{tabular}
\end{center}
where without loss of generality, suppose $\alpha n$ is a large positive integer for a sufficiently large $n$. 

The global optimum for the instance described above is
\begin{align*} 
&(1 0 \cdots 0), &f (1 0 \cdots 0) = n.
\end{align*}

A local optimum  is   \begin{align*}
&(0 \overset{\alpha n-1}{\overbrace{1\cdots 1}} 0 \cdots 0), &f(0 \overset{\alpha n-1}{\overbrace{1\cdots 1}} 0 \cdots 0)=\alpha n-1.
\end{align*}
The ratio of fitness between the local optimum and the global optimum is
 \begin{align*}
 \frac{\alpha n-1}{n} <\alpha.
 \end{align*}

Suppose that the $(N+1)$ EA starts at the above local optimum having the 2nd highest fitness. Truncation selection combined with the ratio-greedy repair prevents a mutant solution  from entering into the next generation unless the mutant individual is the global optimum itself.
Thus, it arrives at the global optimum only if $\alpha n -1$ one-valued bits are flipped into zero-valued ones and the bit $x_1$ is flipped from $x_i=0$ to $x_i= 1$; other zero-valued bits remain unchanged. The probability of this event happening is  
 \begin{align*}
 \left( \frac{1}{n}\right)^{\alpha n} \left( 1-\frac{1}{n}\right)^{n-\alpha n} .
\end{align*}
Thus, we now deduce that the expected runtime is $\Omega(n^{\alpha n} )$, that is exponential in $n$. This completes the argument.
\end{IEEEproof}

Let the constant $\alpha$ towards $0$, proposition~\ref{thmRatioRepair} tells us that the solution produced by the $(N+1)$ EA using  the ratio-greedy repair after a polynomial runtime may be arbitrarily bad.

Next, we consider another pure strategy $(N+1)$ EA that uses the random-greedy repair to tackle the 0-1 knapsack problem, which is described  as follows.

 \begin{algorithmic} [1]
\STATE \textbf{input} an instance of the 0-1 knapsack problem;
\STATE initialize a population considering of N individuals;
\FOR{$t=0, 1,2, \cdots$}
\STATE    mutate  one individual and generate a child;
\IF{the child  is an infeasible solution}
\STATE repair it into a feasible solution using the random-greedy repair;
\ENDIF
\STATE  select N individuals from the parent population and the child using   truncation selection;
 
\ENDFOR
\STATE \textbf{output} the maximum of the fitness function.
\end{algorithmic}

Similarly, we may prove that this EA cannot produce a good solution to the 0-1 knapsack problem within a polynomial runtime using the same instance as that in Proposition~\ref{thmRatioRepair}.
\begin{proposition}
\label{thmRandomRepair}
For any constant $\alpha \in (0,1)$,  the $(N+1)$ EA using Random Repair is not an $\alpha$-approximation algorithm for the 0-1 knapsack problem.
\end{proposition} 

Proposition~\ref{thmRandomRepair} tells us that the solution produced by the $(N+1)$ EA using   random repair is arbitrary bad.

Finally we investigate a pure strategy $(N+1)$ EA using profit-greedy repair for solving the 0-1 knapsack problem, which  is described as follows.
 
 \begin{algorithmic} [1]
\STATE \textbf{input} an instance of the 0-1 knapsack problem;
\STATE initialize a population considering of N individuals;
\FOR{$t=0, 1,2, \cdots$}
\STATE    mutate  one individual and generate a child;
\IF{the child  is an infeasible solution}
\STATE repair it into a feasible solution using the profit-greedy repair;
\ENDIF
\STATE  select N individuals from the parent population and the child using   truncation selection;
 
\ENDFOR
\STATE \textbf{output} the maximum of the fitness function.
\end{algorithmic}

\begin{proposition}
\label{thmProfitRepair}
For any constant $\alpha \in (0,1)$, the $(N+1)$ EA using  profit-greedy repair is not an $\alpha$-approximation algorithm for the 0-1 knapsack problem.
\end{proposition}

\begin{IEEEproof}
Let's consider the following instance:
\begin{center}
\begin{tabular}{c|c|c }
\toprule
Item $i$ & $1$ &   $2, \cdots,  n$ \\
\midrule
Profit $p_i$ &  $\alpha (n-1)$ &$1 $  \\
Weight $w_i$ &  $n-1$  &  $1$     \\
\midrule
Capacity &\multicolumn{2}{c}{$n  $}  \\
\bottomrule
\end{tabular}
\end{center}
 where without loss of generality, suppose $\alpha (n-1)$ is a large positive integer for a sufficiently large $n$.

The local optimum is
\begin{align*}
&(1 0 \cdots 0), &f (1 0 \cdots 0) = \alpha (n-1).
\end{align*}
and the global optimum  is \begin{align*}
&(0 \overset{  n}{\overbrace{1 \cdots 1}}  ), &f(0 \overset{  n}{\overbrace{1 \cdots  1}}  )= n-1.
\end{align*}

The fitness ratio between the local optimum and the global optimum is
 \begin{align*}
 \frac{\alpha (n-1)-1}{n-1} <\alpha.
 \end{align*}

Suppose that the $(N+1)$  EA starts at the local optimum $(1 0  \cdots  0)$.
Let's investigate the following mutually exclusive and exhaustive events:
\begin{enumerate}
\item An infeasible solution has been generated. In this case the infeasible solution will be repaired back to $(1  0  \cdots  0)$ by profit-greedy repair.

\item A feasible solution having the fitness smaller than $\alpha (n -1)$ has been generated. In this case, truncation selection will prevent the new feasible solution from being accepted.

\item A feasible solution is generated having fitness not smaller than $\alpha (n -1)$. This is the only way in which truncation selection will preserve the new mutant solution. Nonetheless, this event happens only if the first bit of the individual in the initial population, $\Phi_0$, is flipped from $x_1 = 1$ into $x_1 = 0$ while at least $\alpha (n-1)$ zero-valued bits of this individual, are flipped from $x_i=0$ into $x_i=1$. The probability of this event is 
 \begin{align*}
 &\sum^n_{k=\alpha (n-1)}  \frac{1}{n} \binom{n}{k}  \left( \frac{1}{n}\right)^{k} \left( 1-\frac{1}{n}\right)^{ n-1-k} \\
 & \le
 O\left( \frac{e}{\alpha (n-1)}\right)^{\alpha (n-1)}.
 \end{align*}
\end{enumerate}

It follows immediately   that if the EA  starts at the local optimum $(1  0  \cdots  0 )$, the expected runtime to produce a better solution    is exponential in $n$. The desired conclusion now follows immediately from definition~\ref{alphaApproxDefn}.
\end{IEEEproof}

Proposition~\ref{thmProfitRepair} tells us that a solution produced by the $(N+1)$ EA using profit-greedy repair may be arbitrarily bad as well.

In summary, we have demonstrated that none of the three pure strategy $(N+1)$ EAs is an $\alpha$-approximation algorithm for the 0-1 knapsack problem given any constant $\alpha \in (0,1).$

\section{Mixed Strategy (N+1) Evolutionary Algorithm}
\label{secMixed}
In this section we analyse mixed strategy  evolutionary algorithm which combines several repair methods together. Here a \emph{mixed strategy EA} refers to an EA employing two or more repairing methods selected with respect to a probability distribution over the set of repairing methods. It may be worth noting that other types of mixed strategy EAs  have been considered in the literature. For example,the mixed strategy EA in~\cite{he2013mixed} employs four mutation operators.  Naturally, we want to know whether or not a mixed strategy (1+1) EA, combining two or more repair methods together, may produce an approximation solution with a guarantee to the 0-1 knapsack problem.

A mixed strategy (1+1) EA for solving the 0-1 knapsack problem is described as follows. The EA combines both, ratio-greedy and profit-greedy repair methods together.
  
 \begin{algorithmic} [1]
\STATE \textbf{input} an instance of the 0-1 knapsack problem;
\STATE initialize a population considering of N individuals;
\FOR{$t=0, 1,2, \cdots$}
\STATE    mutate  one individual and generate a child;
\IF{the child  is an infeasible solution}
  \STATE select either  ratio-greedy repair or  profit-greedy repair method uniformly at random;
\STATE repair it into a feasible solution;
\ENDIF
\STATE  select N individuals from the parent population and the child using   truncation selection;
 
\ENDFOR
\STATE \textbf{output} the maximum of the fitness function.
\end{algorithmic}

Unfortunately the quality of solutions in the mixed strategy EA still has no guarantee.

 \begin{proposition}
\label{thmMixedEA}
Given any constant $\alpha \in (0,1)$, the mixed strategy $(N+1)$ EA  using  ratio-greedy repair and   profit-greedy Repair is not an  $\alpha$-approximation algorithm for the 0-1 knapsack problem.
\end{proposition}

\begin{IEEEproof}
consider the same instance as that in the proof of Proposition~\ref{thmProfitRepair}:
\begin{center}
\begin{tabular}{c|c|c }
\toprule
Item $i$ & $1$ &   $2, \cdots,  n$ \\
\midrule
Profit $p_i$ &  $\alpha (n-1)$ &$1 $  \\
Weight $w_i$ &  $n-1$  &  $1$     \\
\midrule
Capacity &\multicolumn{2}{c}{$n  $}  \\
\bottomrule
\end{tabular}
\end{center}
where the local optimum is
$(1  0  \cdots   0) $ and $f (1 0 \cdots 0) = \alpha (n-1)$.
The global optimum  is $(0 1\cdots 1  )$ and $f(01\cdots 1 )= n-1$.

The fitness ratio between the local optimum and the global optimum is
 \begin{align*}
 \frac{\alpha (n-1)-1}{n-1} <\alpha.
 \end{align*}

Suppose the $(N+1)$ EA starts at the local optimum $(1 0  \cdots 0) $. Let's analyse the following mutually exclusive and exhaustive events that occur upon completion of mutation:
\begin{enumerate}
\item A feasible solution is generated the fitness of which is smaller than $\alpha (n -1)$. In this case, truncation selection will prevent the new feasible solution from entering the next generation.

\item A feasible solution is generated the fitness of which is not smaller than $\alpha (n -1)$. The truncation selection may allow the new feasible solution to enter the next generation.  This event happens only if the first bit is flipped from $x_1=1$ to $x_1=0$ and at least $\alpha (n-1)$ zero-valued bits are flipped into one-valued. The probability of the event is then
is  \begin{align*}
 O\left( \frac{e}{\alpha (n-1)}\right)^{\alpha (n-1)}.
 \end{align*}

\item An infeasible solution is generated, but fewer than $\alpha (n-1)$ zero-valued bits are flipped into the one-valued bits. In this case, either the infeasible solution will be repaired into $(1  0  \cdots  0)$ through the profit-greedy repair; or, it is repaired into a feasible solution where $x_0=0$ and fewer than  $\alpha (n-1)$ one-valued bits among the rest of the bits through the ratio-greedy repair. In the later case the fitness of the  new feasible solution is smaller than $\alpha (n-1)$ and, therefore, cannot be accepted by the truncation selection.

\item An infeasible solution is generated but no fewer than $\alpha (n-1)$ zero-valued bits are flipped into the one-valued bits.
This event happens only if at least $\alpha (n-1)$ zero-valued bits are flipped into the one-valued bits. The probability of the event is then
is  \begin{align*}
 O\left( \frac{e}{\alpha (n-1)}\right)^{\alpha (n-1)}.
 \end{align*}

Afterwards, with a positive probability, it is repaired into a feasible solution where $x_0=0$ and fewer than  $\alpha (n-1)$ one-valued bits among the rest of the bits by the ratio-greedy repair. In the later case the fitness of the  new feasible solution is smaller than $\alpha (n-1)$ and, therefore, it is prevented from entering the next generation by the truncation selection.
\end{enumerate}

Summarizing the four cases described above, we see that when the EA starts at the local optimum $(1  0  \cdots 0)$, it is possible to generate a better solution with probability
is 
 \begin{align*}
 O\left( \frac{e}{\alpha (n-1)}\right)^{\alpha (n-1)}.
 \end{align*}

We then know that the expected runtime to produce a better solution   is exponential in $n$. The conclusion of proposition~\ref{thmMixedEA} now follows at once.
\end{IEEEproof}

Proposition~\ref{thmMixedEA} above tells us that solutions produced by the mixed strategy (M+1) EA exploiting the ratio-greedy repair and  profit-greedy repair may be arbitrarily bad. 

Furthermore, we can prove, that   even the mixed strategy $(N+1)$ EA combining the ratio-greedy repair,  profit-greedy repair and random-repair together, is not an $\alpha$-approximation algorithm for the 0-1 knapsack problem. Its proof is practically identical to that of Proposition~\ref{thmMixedEA}.
 
In summary, we have demonstrated that mixed strategy $(N+1)$ EAs are  $\alpha$-approximation algorithms for the 0-1 knapsack problem given any constant $\alpha \in (0,1).$

\section{Multi-Objective Evolutionary Algorithm}
\label{secMOEA}
So far, we have established several negative results about   $(N+1)$ EAs for the 0-1 knapsack problem.   A naturally arising important question is then how  we can construct an evolutionary approximation algorithm. The most straightforward approach is to apply an approximation algorithm first to produce a good solution, and, afterwards, to run an EA to seek the global optimum solution. Nonetheless, such EAs sometimes get  trapped into the absorbing area of a local optimum,   so it is less efficient   in seeking the global optimum.  

Here we analyse a multi-objective EA using helper objectives (denoted  by MOEA in short), which is similar to the EA presented in~\cite{he2014novel}, but small changes are made in helper objectives for the sake of analysis. Experiment results in~\cite{he2014novel} have shown that the MOEA using helper objectives  performs better than the simple combination of an approximation algorithm and a GA.

The MOEA is designed using the \emph{multi-objectivization} technique.  In multi-objectivization, single-objective optimisation problems are transferred  into multi-objective optimisation problems by decomposing the original objective into several components~\cite{knowles2001reducing} or by adding  helper objectives~\cite{jensen2005helper}. Multi-objectivization may bring both positive and  negative  effects~\cite{handl2008multiobjectivization,brockhoff2009effects,lochtefeld2011helper}.   This approach has been used for solving several combinatorial optimisation problems,  for example, the  knapsack problem \cite{kumar2006analysis}, vertex cover problem \cite{friedrich2010approximating} and minimum label spanning tree problem \cite{lai2014performance}.

Now we describe the MOEA using helper objectives, similar to the EA in~\cite{he2014novel}. The original single objective optimization problem (\ref{equ:single-objective}) is recast into a multi-objective optimization problem using three helper objectives.    
First let's look at the following instance. 
\begin{center}
\begin{tabular}{c|c|c|c|c|c}
\toprule
Item  & 1  & 2 & 3 & 4 &5 \\
\midrule
Profit   &  10 & 10  &  10 & 12  & 12\\
Weight   & 10  & 10 & 10   & 10  &10 \\
\midrule
Capacity &\multicolumn{5}{c}{20}\\
\bottomrule
\end{tabular}
\end{center}

The global optimum  is  
 $00011$ in this instance.   In the optimal solution, the  average profit of packed items is the largest. Thus the first helper objective is to maximize the average profit of items in a knapsack.  We don't use the original value of profits, instead we use the ranking value of profits. Assume that the profit of item $i$ is the $k$th smallest, then let the ranking value $\hat{p}_i=k$. For example in the above instance,  $\hat{p}_1=\hat{p}_2=\hat{p}_3=1$ and $\hat{p}_4=\hat{p}_5=2$. Then the  helper objective function is defined to be
\begin{equation}
   h_1(\vec{x}) = \frac{1}{\parallel \vec{x} \parallel_1}\sum^n_{i=1}   x_i \hat{p}_i,
\end{equation}
where $\parallel \vec{x} \parallel_1 =\sum^n_{i=1} x_i$.

Next we consider another instance.
\begin{center}
\begin{tabular}{c|c|c|c|c|c}
\toprule
Item  & 1  & 2 & 3 & 4 &5 \\
\midrule
Profit   &  15 & 15  &  20 & 20  & 20\\
Weight   & 10  & 10 & 20   & 20  &20 \\
\midrule
Capacity &\multicolumn{5}{c}{20}\\
\bottomrule
\end{tabular}
\end{center}

The global optimum  is  
 $11000$  in this instance.   In the optimal solution, the  average  profit-to-weight ratio of packed items is the largest. However, the average profit of these items is not the largest. Then  the second helper objective is to maximize the average profit-to-weight ratio of items in a knapsack.   We don't use the original value of profit-to-weight, instead  its ranking value. Assume that the profit-to-weight of item $i$ is the $k$th smallest, then let the ranking value $\hat{r}_i=k$. For example in the above instance,  $\hat{r}_1=\hat{r}_2=2$ and $\hat{r}_3=\hat{r}_4=\hat{r}_5=1$. Then the  helper objective function is defined to be  
\begin{equation}
   h_2(\vec{x}) = \frac{1}{\parallel \vec{x} \parallel_1}\sum^n_{i=1}   x_i \hat{r}_i.
\end{equation} 

Finally let's see the following instance.
\begin{center}
\begin{tabular}{c|c|c|c|c|c}
\toprule
Item  & 1  & 2 & 3 & 4 &5 \\
\midrule
Profit   &  40 & 40  &  40 & 40 & 150\\
Weight   & 30 & 30&30   & 30  &100 \\
\midrule
Capacity &\multicolumn{5}{c}{120}\\
\bottomrule
\end{tabular}
\end{center}
 
The global optimum is  
 $11110$ in this instance.   In the optimal solution,  neither the average profit of packed items nor average profit-to-weight ratio  is  the largest. Instead the  number of packed items is the largest, or the average weight is the smallest. Thus the  third helper objectives are to maximize the number of items in a knapsack.    The objective functions   are
\begin{eqnarray}
  & h_3(\vec{x}) = \parallel \vec{x} \parallel_1. 
\end{eqnarray}

 We then consider a multi-objective optimization problem:
\begin{equation}
\label{equ:multi-objective}
\begin{array}{lll}
 \max_{\vec{x}} \{ f(\vec{x}),  h_1(\vec{x}) , h_2(\vec{x}),   h_3(\vec{x}) \},
 \\ \mbox{subject to } \sum^n_{i=1} w_i x_i \le C.
\end{array}
\end{equation}

The multi-objective optimisation problem (\ref{equ:multi-objective})  is solved by an EA using  bitwise mutation,  and multi-criteria truncation selection, plus a mixed strategy of two repair methods.  
 \begin{algorithmic} [1]
\STATE \textbf{input} an instance of the 0-1 knapsack problem;
\STATE initialize a population considering of N individuals;
\FOR{$t=0, 1,2, \cdots$}
\STATE    mutate  one individual and generate a child;
\IF{the child  is an infeasible solution}
  \STATE select either  ratio-greedy repair or  profit-greedy repair method uniformly at random;
\STATE repair it into a feasible solution;
\ENDIF
\STATE  select N individuals from the parent population and the child using the multi-criterion truncation selection;
 
\ENDFOR
\STATE \textbf{output} the maximum of the fitness function.
\end{algorithmic}

A novel \emph{multi-criteria truncation selection operator} is  adopted in the above EA. Since the target is to maximise several objectives simultaneously, we  select a few individuals which have higher function values with respect to each objective function.  The pseudo-code of multi-criteria selection is described as follows.

\begin{algorithmic} [1]
\STATE \textbf{input} the parent population   and  the child;
 
\STATE merge the parent  population and the child into a temporary population   which consists of $N+1$ individuals;

\STATE sort  all individuals in  the temporary population in the descending order of $f(\vec{x})$, denote them by $\vec{x}^{(1)}_1, \cdots, \vec{x}^{(1)}_{N+1}$; 
 
\STATE select all individuals from left to right (denote them by $\vec{x}^{(1)}_{k_1}, \cdots, \vec{x}^{(1)}_{k_m}$) which satisfy $h_1(\vec{x}^{(1)}_{k_i})< h_1(\vec{x}^{(1)}_{k_{i+1}})$ or $h_2(\vec{x}^{(1)}_{k_i})< h_2(\vec{x}^{(1)}_{k_{i+1}})$  for any $k_i$. 
\IF{the   number of selected individuals  is greater than $\frac{N}{3}$}
\STATE truncate them to $\frac{N}{3}$ individuals;
\ENDIF

\STATE add the  selected individuals into the next generation population;

\STATE resort  all individuals in  the temporary population  in the descending order of  $h_1(\vec{x})$,  still denote them by $\vec{x}_1, \cdots, \vec{x}_{N+1}$;    

\STATE select all individuals from left to right (still denote them by $\vec{x}_{k_1}, \cdots, \vec{x}_{k_m}$) which satisfy $h_3(\vec{x}_{k_i})< h_3(\vec{x}_{k_{i+1}})$  for any $k_i$. 
\IF{ the   number of selected individuals is greater than $\frac{N}{3}$}
\STATE truncate them to $\frac{N}{3}$ individuals;
\ENDIF
\STATE add the  selected individuals into the next generation population; 

\STATE resort  all individuals in  the temporary population  in the descending order of $h_2(\vec{x})$,  still denote them by $\vec{x}_1, \cdots, \vec{x}_{N+1}$; 
 
\STATE select all individuals from   left to right (still denote them by $\vec{x}_{k_1}, \cdots, \vec{x}_{k_m}$) which satisfy $h_3(\vec{x}_{k_i})< h_3(\vec{x}_{k_{i+1}})$ for any $k_i$. 
\IF{ the   number of selected individuals is greater than  $\frac{N}{3}$}
\STATE truncate them to $\frac{N}{3}$ individuals;
\ENDIF
\STATE add these selected individuals into the next generation population; 

\WHILE{the  next generation population size  is less than  $N$} 
\STATE randomly choose an individual from the  parent population and child,  and add it into the next generation population;
\ENDWHILE

\STATE \textbf{output} a new population $\Phi_{t+1}$.
\end{algorithmic}

In the above algorithm, Steps 3-4 are for selecting the individuals  with higher values of $f(\vec{x})$. In order to preserve  diversity, we  choose these individuals which have different values of $h_1(\vec{x})$ or $h_2(\vec{x})$. Similarly
Steps 9-10 are for selecting the individuals with a higher value of $h_1(\vec{x})$. We  choose the individuals which have different values of $h_3(\vec{x})$ for maintaining diversity.
Steps 15-16 are for selecting individuals with a higher value of $h_{2}(\vec{x})$. Again we  choose these individuals which have different values of $h_3(\vec{x})$ for preserving diversity. 
We don't explicitly select individuals based on  $h_3(\vec{x})$. Instead we implicitly do it  during  Steps 9-10, and Steps 15-16.

Using helper objectives and multi-criterion truncation selection brings a benefit of  searching along several directions $f(\vec{x}), h_1(\vec{x})$, $h_2(\vec{x})$ and implicitly $h_3(\vec{x})$. Hence the MOEA may arrive at a local optimum quickly, but at the same time, does not get trapped into the absorbing area of a local optimum of $f(\vec{x})$. The experiment results~\cite{he2014novel} have demonstrate the MOEA using helper objectives outperform the simplified combination of an approximation algorithm and a  GA.

 The analysis is based on a  fact which is derived from the analysis of the greedy algorithm for the 0-1 knapsack problem (see \cite[Section 2.4]{martello1990knapsack})). Consider  the following   algorithm:
\begin{algorithmic} [1]
\STATE  let   $\vec{a}^*$ be the feasible solutions with the largest profit item;

 \STATE resort all the items via the ratio of their profits to their corresponding weights so that $\frac{p_1}{w_1} \ge \cdots \ge \frac{p_n}{w_n}$;
 
\STATE greedily add the items in the above order to the knapsack as long as adding an item to the knapsack does not exceeding the capacity of the knapsack. Denote the solution by $\vec{b}^*$.
\end{algorithmic} 

Then the fitness   of $\vec{a}^*$  or $\vec{b}^*$  is not smaller than  1/2 of the fitness of the optimal solution.

Based on the above fact, we can prove the following result.

\begin{theorem} 
If $N \ge 3n$, then the  $(N+1)$ MOEA can produce a feasible solution, which is not worse than  $\vec{b}^*$ and $\vec{a}^*$, within   $O(N n^3)$ runtime.
\end{theorem}

\begin{IEEEproof}
(1) Without loss of generality, let the first item be the most profitable one.
First,  it suffices to prove that the EA can generate a feasible solution fitting the Holland schema $(1 *\cdots *)$  (as usual, $*$ stands for the `don't care' symbol that could be replaced either by a $1$ or a $0$) within a polynomial runtime.

Suppose that the value of $h_1$ of all the individuals in the population  are smaller than that of $\vec{a}^*$, that is, they fit the Holland schema $(0 * \cdots *)$. Let $\vec{x}$ be the individual that is chosen for mutation. Through mutation, $x_1$ can be flipped from $x_1=0$ to $x_1=1$ with probability $1/n$. If the child is feasible, then we arrive at the desired individual (denote it by $\vec{y}$). If the child is infeasible, then, with probability $1/2$, the first item will be kept thanks to the profit-greedy repair and a feasible solution is generated (denote it by $\vec{y}$). We have now shown that the EA can generate a feasible solution that includes the most profitable item with probability at least $1/(2n)$.  

Thus, the EA can generate a feasible solution fitting the Holland schema $(1 * \cdots *)$ within  the expected runtime is at most $2n$.

(2)  Without loss of generality, let
 \begin{align*}
\frac{p_1}{w_1}> \cdots >  \frac{p_m}{w_m}  > \cdots >  \frac{p_n}{w_n}.
\end{align*}
and let $ \vec{b}^*=  (\overset{m}{\overbrace{1\cdots 1} }0 \cdots 0)$.  We now demonstrate that the EA can reach $\vec{b}^*$ within a polynomial runtime via  objectives $h_2$ and $h_3$. 

First we prove that the EA    can reach $(10 \cdots 0)$ within a polynomial runtime. We exploit drift  analysis~\cite{he2001drift} as a tool to establish the result. For a binary vector $\vec{x}=(x_1  \cdots  x_n)$, define the distance function
\begin{align}
d(\vec{x}) =h_2(10 \cdots 0) -h_2(\vec{x}).
\end{align}
 For a population $(\vec{x}_1, \cdots, \vec{x}_N)$,  its distance function is 
$$
\min \{ d(\vec{x}_1), \cdots, d(\vec{x}_N).\}
$$
According to the definition of $h_2(\vec{x})$, the above distance function is upper-bounded by $n$.
 
Suppose that none of individuals in the current population is $(10 \cdots 0)$. Let  $\vec{x}$ be the individual, the value of whose distance is the smallest in the current population. The individual belongs to one of the two cases below:
\begin{IEEEdescription}
\item [Case 1]: $\vec{x}$ fits the Holland schema $(1*\cdots*)$ where at least one * bit takes the value of 1.

\item [Case 2]: $\vec{x}$ fits the Holland schema $(0*\cdots*)$.
\end{IEEEdescription}

The   individual will be chosen for mutation with probability $\frac{1}{N}$. Now we analyse the mutation event related to the above two cases.

\textbf{Analysis of Case 1}: one of 1-valued *-bits (but not the first bit) is flipped into 0-valued; other bits are not changed. This event will happen with a probability
\begin{align}
\frac{1}{n} \left(1-\frac{1}{n}\right)^{n-1}=\Omega(n^{-1}).
\end{align}

Let's establish how the  value of $h_2$   increases during the mutation. Denote the 1-valued bits in $\vec{x}$ by $i_1, \cdots, i_k$. Then the objective $h_2$'s  value is
$$
\frac{\hat{p}_{i_1}+\cdots + \hat{p}_{i_k}}{k}.
$$ 
Without loss of generality, the $i_k$th bit is flipped into 0-valued. Then after mutation, the 1-valued bits in $\vec{x}$ becomes $i_1, \cdots, i_{k-1}$ and  the objective $h_2$'s value is
$$
\frac{\hat{p}_{i_1}+\cdots + \hat{p}_{i_{k-1}}}{k-1}.
$$ 
Thus, the value of $h_2$ increases (or equivalently, the value of $d$ decreases) by 
\begin{align}
\frac{\hat{p}_{i_1}+\cdots + \hat{p}_{i_{k-1}}}{k-1}- \frac{\hat{p}_{i_1}+\cdots + \hat{p}_{i_k}}{k} =\Omega (n^{-2}).
\end{align}

Thanks to the multi-criteria truncation selection, the value of $h_2$ always increases. So there is no negative drift. Therefore the drift in Case 1 is
\begin{align}
 \Omega (N^{-1} n^{-3} ).
\end{align}

\textbf{Analysis of Case 2}: The first bit is flipped into 0-valued; other bits are not changed. The analysis then is identical to Case 1. The drift in Case 2 is  
$
 \Omega (N^{-1} n^{-3} ) 
$, the same as that in Case 1.

Recall that the distance function $d(\vec{x}) \le n$. Applying the drift theorem~\cite[Theorem 1]{he2001drift}, we deduce that the expected runtime to reach  $(10 \cdots 0)$ is $ O(Nn^3) $.  Once  $(10 \cdots 0)$ is included in the population, it will be kept for ever according to the   multi-criteria truncation selection. 

Next we prove that the EA  can reach $\vec{b}^*$ within a polynomial runtime when starting from $(10 \cdots 0)$. Suppose  that the current population includes an individual $(10\cdots 0)$ but no individual $(110\cdots 0)$. The individual $(10\cdots 0)$ may be chosen for mutation with a probability $\frac{1}{N}$, then it can be mutated into $(110\cdots 0)$ with a probability $\Omega(n^{-1})$. The individual $(110\cdots 0)$ has the second largest value of $h_2$, thus, according to the multi-criteria truncation selection, it will be kept in the next generation population. Hence the expected runtime for the EA to reach the individual $(110\cdots 0)$ is $O(Nn)$. Similarly we can prove that the EA will reach $(1110 \cdots 0)$ within $O(N n)$ runtime, then $(11110\cdots 0)$ within $O(N n)$ runtime, and so on. The expected runtime for the EA to reach $\vec{b}^*$ is $O(Nn^2)$. 

Combining the above discussions together, we see that the expected runtime to produce a solution not worse than $\vec{a}^*$ and $\vec{b}^*$ is $O(N n^3)+O(N n^2)$.     
\end{IEEEproof}

If we change  helper objective functions  $h_1(\vec{x})$ and $h_2(\vec{x})$ to those used in~\cite{he2014novel}, 
\begin{align}
   h_1(\vec{x}) &= \frac{1}{\parallel \vec{x} \parallel_1}\sum^n_{i=1}   x_i {p}_i,\\
   h_2(\vec{x}) &= \frac{1}{\parallel \vec{x} \parallel_1}\sum^n_{i=1}   x_i \frac{p_i}{w_i},
\end{align}  
then  the above proof doesn't work, and we need a new   proof for obtaining the same conclusion. Furthermore, it should be mentioned that none of the three objectives can be removed; otherwise the MOEA will not produce a solution with a guaranteed approximation ratio. But on the other side, the performance might be better if adding  more objectives, for example,
\begin{equation}
\min h_4(\vec{x}) =   \frac{1}{\parallel \vec{x} \parallel_1}\sum^n_{i=1}   x_i  w_i.
\end{equation}

\section{Conclusions}
\label{secConclusions}

In this work, we have assessed  the solution quality in three types of $(N+1)$ EAs, which exploit bitwise mutation and truncation selection, for solving the knapsack problem. We have proven that the pure strategy EAs using a single repair method and the mixed strategy EA combing two repair methods are not a   $\alpha$-approximation algorithm for any constant $\alpha \in (0,1)$. In other words, solution quality in these EAs may be arbitrarily bad. Nevertheless,  we have shown that a multi-objective $(N+1)$ EA using helper objectives    is a 1/2-approximation algorithm. Its runtime is $O(N n^3)$. Our work demonstrates that using  helper objectives is a good approach to design evolutionary approximation algorithms.  The advantages of the EA using helper objectives is to search along several directions and also to preserve population diversity.

Population-based EAs using other strategies of preserving diversity, such as   niching methods,   are not investigated in this paper. The extension of this  work to such EAs will be the future research.  Another work in the future  is to study the solution quality of MOEAs for  the multi-dimension knapsack problem.

\paragraph*{Acknowledgements}
This work was supported by the EPSRC under Grant No. EP/I009809/1 and  by the NSFC  under Grant No. 61170081.
 

\end{document}